\documentclass{article}

\usepackage[nonatbib,preprint]{neurips_2023_modified}

\usepackage[numbers,sort&compress]{natbib}

\usepackage{graphicx}
\usepackage{multirow}
\usepackage{tikz}
\usepackage{caption}
\usepackage{subcaption}
\usepackage{tcolorbox}
\usepackage{diagbox}
\usepackage[inline]{enumitem}
\usepackage{adjustbox}
\usepackage{soul}
\usepackage{amsmath}
\usepackage{amssymb}
\usepackage{booktabs}
\usepackage{hyperref}

\DeclareRobustCommand{\hlred}[1]{{\sethlcolor{red!12}\hl{#1}}}

\usepackage{listings} 
\usepackage{xcolor} 
\definecolor{titlegray}{rgb}{0.9,0.9,0.9} 
\definecolor{ForestGreen}{RGB}{34,139,34}
\definecolor{orange-red}{rgb}{1.0, 0.27, 0.0}
\definecolor{cadmiumred}{rgb}{0.89, 0.0, 0.13}
\definecolor{placeholder}{RGB}{0,120,215}
\definecolor{orangecircle}{rgb}{0.93, 0.53, 0.18}
\definecolor{bluecircle}{rgb}{0.35, 0.31, 0.81}

\lstdefinestyle{python}{
  language=Python,
  basicstyle=\ttfamily\footnotesize,
  keywordstyle=\color{blue},
  stringstyle=\color{purple},
  commentstyle=\color{ForestGreen},
  showstringspaces=false,
  numberstyle=\tiny,
  backgroundcolor=\color{gray!10},
  frame=none, 
  aboveskip=-1pt, 
  belowskip=-1pt,
  lineskip=-1pt, 
  escapeinside={(*@}{@*)},
}

\newcommand{\T}{\ensuremath{\mathcal{T}}}
\newcommand{\Tdesc}{\ensuremath{\mathcal{T}_{\desc}}}
\newcommand{\Ttests}{\ensuremath{\mathcal{T}_{\tests}}}
\newcommand{\desc}{\mathrm{desc}}
\newcommand{\theme}{\mathrm{theme}}
\newcommand{\concepts}{\mathrm{concepts}}
\newcommand{\tests}{\mathrm{tests}}

\newcommand{\GenBase}{\ensuremath{\textsc{Base}}}
\newcommand{\GenConsistency}{\ensuremath{\textsc{GenConsistency}}}
\newcommand{\LLMJudge}{\ensuremath{\textsc{LLMJudge}}}
\newcommand{\SimTutorVal}{\ensuremath{\textsc{SimTutorVal}}}
\newcommand{\SimStudentVal}{\ensuremath{\textsc{SimStudentVal}}}
\newcommand{\PyTaskSyn}{\ensuremath{\textsc{PyTaskSyn}}}
\newcommand{\Oracle}{\ensuremath{\textsc{Oracle}}}

\newcommand{\tickicon}{\includegraphics[height=1.2em]{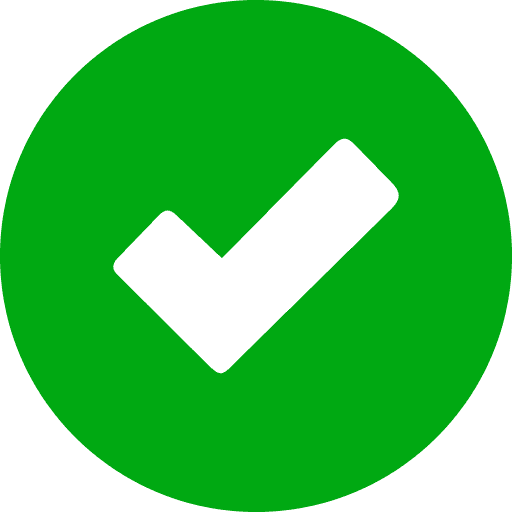}}
\newcommand{\crossicon}{\includegraphics[height=1.2em]{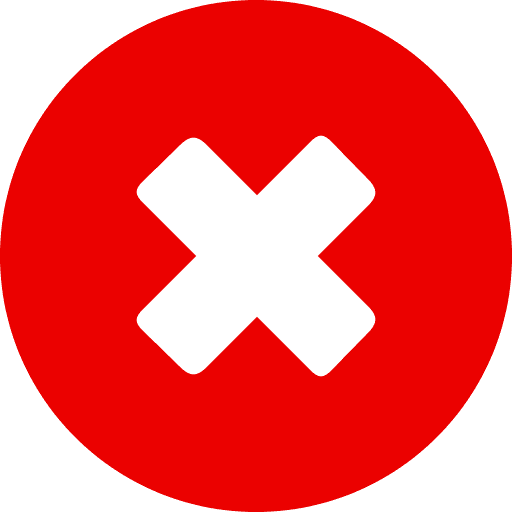}}

\newcommand{\orangecircled}[1]{%
    \tikz[baseline=(X.base)] 
    \node[draw, circle, fill=orangecircle, minimum size=10pt, inner sep=0pt] (X) {\textcolor{white}{\scalebox{0.6}{\textbf{\sffamily{#1}}}}};%
}

\newcommand{\bluecircled}[1]{%
    \tikz[baseline=(X.base)] 
    \node[draw, circle, fill=bluecircle, minimum size=10pt, inner sep=0pt] (X) {\textcolor{white}{\scalebox{0.6}{\textbf{\sffamily{#1}}}}};%
}

\title{Synthesizing High-Quality Programming Tasks \\with LLM-based Expert and Student Agents}
\author{%
  Manh Hung Nguyen\\
  MPI-SWS, Germany\\
  \text{manguyen@mpi-sws.org} \\
  \And
  Victor-Alexandru Pădurean\\ 
  MPI-SWS, Germany\\
  \text{vpadurea@mpi-sws.org} \\
  \And
  Alkis Gotovos\\ 
  MPI-SWS, Germany\\
  \text{agkotovo@mpi-sws.org} \\
  \And
  Sebastian Tschiatschek \\
  University of Vienna, Austria \\
  \text{sebastian.tschiatschek@univie.ac.at} \\
  \And
  Adish Singla \\
  MPI-SWS, Germany \\
  \text{adishs@mpi-sws.org} \\
}

\begin{document}

\maketitle


\begin{abstract}
Generative AI is transforming computing education by enabling the automatic generation of personalized content and feedback. We investigate its capabilities in providing high-quality programming tasks to students. Despite promising advancements in task generation, a quality gap remains between AI-generated and expert-created tasks. The AI-generated tasks may not align with target programming concepts, could be incomprehensible to students, or may contain critical issues such as incorrect tests. Existing works often require interventions from human teachers for validation. We address these challenges by introducing \PyTaskSyn{}, a novel synthesis technique that first generates a programming task and then decides whether it meets certain quality criteria to be given to students. The key idea is to break this process into multiple stages performed by expert and student agents simulated using both strong and weaker generative models. Through extensive evaluation, we show that \PyTaskSyn{} significantly improves task quality compared to baseline techniques and showcases the importance of each specialized agent type in our validation pipeline. Additionally, we conducted user studies using our publicly available web application and show that \PyTaskSyn{} can deliver high-quality programming tasks comparable to expert-designed ones while reducing workload and costs, and being more engaging than programming tasks that are available in online resources.

\end{abstract}

\section{Introduction}
Generative AI is transforming learning and teaching in computing education \cite{DBLP:journals/corr/abs-2402-01580,DBLP:conf/iticse/Prather00BACKKK23}. Advanced generative models such as OpenAI's GPT-4o \cite{GPT4o} and GitHub Copilot \cite{CopilotWeb} are quickly reshaping both student and teacher experiences. For students, these models can provide personalized educational content \cite{DBLP:conf/educon/Abu-Rasheed0F24}, provide detailed feedback on their work \cite{leinonen23sigcse,DBLP:conf/lak/PhungPS0CGSS24}, and serve as pair programmers \cite{DBLP:conf/chi/MozannarBFH24}. For teachers, these models can assist in analyzing and grading student answers \cite{DBLP:conf/icer/PhungPCGKMSS22} and curriculum development \cite{DBLP:conf/aied/SridharDABSS23}. A particularly promising application is their ability to generate tailored educational materials, especially in creating diverse programming exercises that target specific concepts.

Recent works have investigated the use of generative models for generating novel and engaging programming exercises related to a specific theme and targeting specific programming concepts \cite{DBLP:conf/iticse/Gutierrez0L24,DBLP:conf/sigcse/JordanLR24,DBLP:conf/icer/LogachevaHPS024,DBLP:conf/icer/SarsaDH022,DBLP:conf/dsaa/KhanRNK24}. While these initial efforts show promise, AI-generated tasks still fall short of human expert quality due to several issues \cite{DBLP:conf/sigcse/JordanLR24,DBLP:conf/icer/PhungPCGKMSS22,DBLP:conf/icer/SarsaDH022}. For example, the generated programming task can contain incorrect test cases generated as part of the task or it can contain a task description that is not comprehensible \cite{DBLP:conf/sigcse/JordanLR24,DBLP:conf/icer/SarsaDH022}. Without an automatic validation mechanism to check these aspects of the task, human interventions would be required to validate the task's quality before they are assigned to students \cite{DBLP:conf/sigcse/Gutierrez0L24,DBLP:conf/sigcse/JordanLR24}. While one might consider using generative models for task validation, research has shown that they struggle with self-correction \cite{DBLP:conf/iclr/0009CMZYSZ24}. These limitations in single-agent validation motivated our multi-agent based approach.

\begin{figure*}[t]
    \centering
        \centering
        \includegraphics[width=1\textwidth, trim={6cm 5cm 7cm 5cm}, clip]{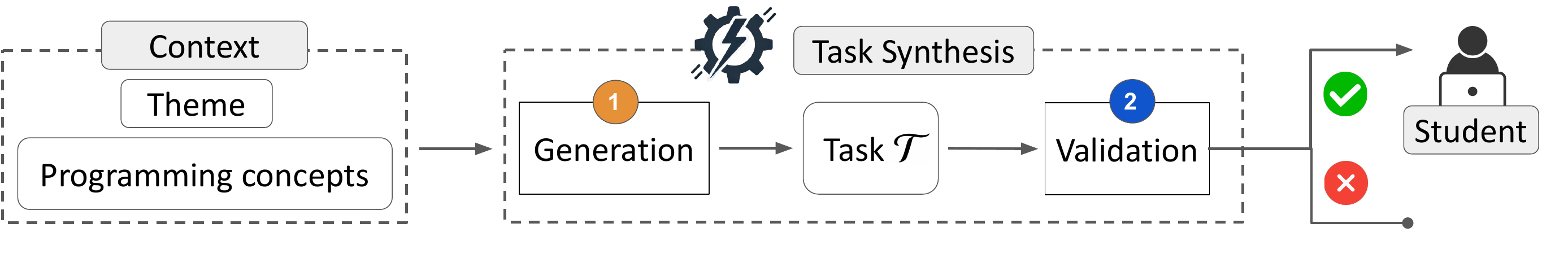}
    \caption{Overview of contextualized programming task synthesis pipeline. Given an input context consisting of a theme and a set of programming concepts, a task \T{} is first generated in Stage \protect\orangecircled{1} and then validated in Stage \protect\bluecircled{2}. Task \T{} is assigned to a student only if it meets certain quality criteria. }     
    \label{fig.intro.overview}
\end{figure*}

To address these challenges, we introduce our technique, \PyTaskSyn{}, capable of generating contextualized programming tasks and then deciding whether they meet certain quality criteria. Figure~\ref{fig.intro.overview} depicts the overview of the pipeline implemented in \PyTaskSyn{}. First, given a context as input, \PyTaskSyn{} asks a simulated agent to generate a programming task composed of a description and a test suite. The generated task then goes through the second stage in the pipeline (i.e., validation), handled by multiple simulated agents with unique roles using strong (GPT-4o \cite{GPT4o}) and weaker (GPT-4o-mini \cite{GPT4omini}) models. This validation stage is designed to provide a quality assurance mechanism and decides whether the generated task can be provided to a student or not. Our multi-agent approach builds upon research that has shown the benefit of collaborative agents \cite{DBLP:conf/iclr/ChenSZ0YCYLHQQC24,DBLP:conf/icml/Du00TM24,DBLP:journals/corr/abs-2308-08155} and used generative models to simulate humans in various roles \cite{DBLP:conf/icml/AherAK23,DBLP:conf/lats/Lu024,DBLP:conf/edm/PhungCGKMSS23,DBLP:conf/lak/PhungPS0CGSS24,2024.EDM-short-papers.31}. We demonstrate the efficacy of our technique through extensive evaluation using data collected from prior works \cite{DBLP:conf/iticse/Gutierrez0L24,DBLP:conf/sigcse/Gutierrez0L24,DBLP:conf/icer/LogachevaHPS024,DBLP:conf/icer/SarsaDH022}.

In summary, our main contributions are as follows.
\begin{enumerate}[label={(\arabic*)},leftmargin=2em]

    \item We highlight the effectiveness of decomposing the programming task synthesis into multiple stages. We introduce \PyTaskSyn{}, a novel technique that leverages generative models to simulate expert, tutor, and student agents, each responsible for specific stages in the synthesis process. (Section \ref{sec.technique}).\footnote{\url{https://github.com/machine-teaching-group/aied2025-pytasksyn}}
    
    \item We conduct extensive evaluation of our technique for Python programming task synthesis, demonstrating significant improvements in the quality of synthesized tasks, while maintaining substantial coverage (Section \ref{sec.evaluationsetup}).
    
    \item We develop a web application for Python programming task synthesis and conduct two user studies, showing that our synthesized tasks match the quality of expert-created tasks while requiring minimal cost. (Section \ref{sec.pilotstudy}).
\end{enumerate}


\section{Related Work}
\textbf{Programming task generation.}
 With the advancement of generative models, recent studies have demonstrated the potential of using them to create natural language programming tasks \cite{DBLP:conf/iticse/Gutierrez0L24,DBLP:conf/sigcse/Gutierrez0L24,DBLP:conf/sigcse/JordanLR24,DBLP:conf/icer/LogachevaHPS024,DBLP:conf/icer/PhungPCGKMSS22,DBLP:conf/icer/SarsaDH022}. Initial research focused on the generation of programming exercises, including sample solutions and test cases \cite{DBLP:conf/icer/SarsaDH022}, while subsequent research efforts have aimed at synthesizing more diverse and contextualized programming exercises \cite{DBLP:conf/iticse/Gutierrez0L24,DBLP:conf/sigcse/Gutierrez0L24,DBLP:conf/icer/LogachevaHPS024}, as well as exercises suitable for non-English contexts \cite{DBLP:conf/sigcse/JordanLR24}. A parallel line of work focused on generating programming tasks to target specific types of bugs \cite{DBLP:conf/icer/PhungPCGKMSS22}, or visual programming tasks \cite{DBLP:conf/nips/AhmedCEFGRS20,padurean2024neural,DBLP:conf/aied/WenGSS24}. However, many of these works highlighted the significant gap between AI-generated tasks and human-crafted tasks \cite{DBLP:conf/sigcse/JordanLR24,padurean2024neural,DBLP:conf/icer/PhungPCGKMSS22,DBLP:conf/icer/SarsaDH022}. Our work addresses this gap by automating the validation of programming tasks' quality.

\textbf{Improving quality of programming content.} 
Recent works have focused on enhancing the quality of programming content. For example, generative models were used to generate predicates for testing a generated program \cite{DBLP:journals/corr/abs-2210-00848}, or to act as judges for evaluation \cite{DBLP:conf/nips/ZhengC00WZL0LXZ23}. As an enhancement, generative models can leverage their own evaluations to iteratively improve the quality of their responses \cite{DBLP:conf/iclr/ChenLSZ24,DBLP:conf/nips/MadaanTGHGW0DPY23}. However, prior work has shown that validating the feedback generated by one generative agent using another generative agent has proved to be more effective than using the same agent \cite{DBLP:conf/edm/PhungCGKMSS23,DBLP:conf/lak/PhungPS0CGSS24}. Similarly, in our work, we break the task synthesis process into smaller processes and assign them to different agents.

\looseness-1\textbf{Generative models as simulated agents.}
A recent line of research involves agents assuming different roles and collaborating to achieve a goal. For example, in AutoGen \cite{DBLP:journals/corr/abs-2308-08155}, multiple agents can converse, use tools, and incorporate human input. Similar studies show that simulated agents interacting with each other outperform a single agent in reasoning and planning \cite{DBLP:conf/iclr/ChenSZ0YCYLHQQC24,DBLP:conf/icml/Du00TM24}. Several studies have explored LLM-empowered agents for simulating classroom interactions \cite{DBLP:journals/corr/abs-2406-19226}, facilitating tutor training \cite{DBLP:conf/lats/MarkelOLP23}, and implementing learning-by-teaching environments \cite{DBLP:conf/chi/JinLSK24,DBLP:conf/aied/MaSKW24,DBLP:journals/corr/abs-2310-01420}. Our work builds upon these successes by simulating experts, tutors, and students while synthesizing tasks to ensure their quality.


\section{Problem Setup}
\label{sec.problemsetup}
\begin{figure}[t]
\footnotesize
\begin{adjustbox}{minipage=[t][0.6\textheight]{1\textwidth},fbox}
    \begin{subfigure}[t]{0.995\textwidth}
      \parbox{\textwidth}{
        \textbf{Theme:} Superheroes}\\
        \textbf{Programming concepts:} \colorbox{red!12}{Dictionaries}, Classes $\&$ Objects, Strings, Arithmetic Operators
      \vspace{0.05cm}

      \colorbox{gray!8}{
        \parbox{\dimexpr\textwidth-2\fboxsep\relax}{
        \footnotesize
        \centerline{\textbf{Task Description}}
        Design and implement a Python class `Superhero' that models a simple superhero character using the following guidelines:
        
        1. Attributes:

            \hspace{0.3cm}- `name' (string) : Name of the superhero

            \hspace{0.3cm}- `power' (string) : A short description of their superpower

            \hspace{0.3cm}- `age' (integer) : Age of the superhero

            \hspace{0.3cm}- `world\_saving\_points' (integer) : Points representing the superhero's achievements.
        
        2. Methods:

        \hspace{0.3cm}- `\_\_init\_\_(self, name, power, age)' : This method should initialize a superhero with the provided name,
        
        \hspace{0.5cm} power, and age. The `world\_saving\_points' should start at 0.
           
        \hspace{0.3cm}- `save\_the\_day(self, difficulty)' : This method takes a difficulty level (integer) and increases the  
        
        \hspace{0.5cm}`world\_saving\_points' by two times the difficulty level. If `difficulty' is less than 1, it should not change 
        
        \hspace{0.5cm}the points.
           
        \hspace{0.3cm}- `get\_description(self)' : This method returns a string describing the superhero in the format: 
        
        \hspace{0.5cm}``\{name\} possesses the power of \{power\} and is \{age\} years old.''
        
        3. Functions:

        \hspace{0.3cm}- Implement a standalone function `top\_hero(hero\_list)' that takes a list of `Superhero' objects and returns
        
        \hspace{0.5cm}the name of the superhero with the most `world\_saving\_points'. If there is a tie, return the 
        
        \hspace{0.5cm}lexicographically smaller name.
        
        }
      }

      \vspace{0.25cm}

\begin{lstlisting}[ style=python,frame=single,framerule=0mm,framexbottommargin=0mm,escapeinside={(*}{*)}]
(*\centerline{\footnotesize\rmfamily\textbf{Test suite}}*) 
def test_top_hero():
  superheroes = [Superhero("Thor","thunder god",1500), Superhero("Hulk","super          
             strength",35), Superhero("Doctor Strange","magic",45)]
  superheroes[0].save_the_day(10)
  superheroes[1].save_the_day(10)
  superheroes[2].save_the_day(12)
  assert top_hero(superheroes) == "Doctor Strange"
  superheroes[1].save_the_day(4)  
  \end{lstlisting}
\begin{lstlisting}[ style=python,backgroundcolor=\color{red!12},frame=single,framerule=0mm,framexbottommargin=1mm,escapeinside={(*}{*)}]
  assert top_hero(superheroes) == "Doctor Strange" 
  \end{lstlisting}

\begin{lstlisting}[ style=python,frame=single,framerule=0mm,framexbottommargin=0mm,escapeinside={(*}{*)}]
(*\rmfamily...(\textit{other test cases are ommited for brevity)}*)
\end{lstlisting}

    \vspace{0.2cm}
      \centering
      \begin{tabular}{>{\centering\arraybackslash}p{0.22\textwidth} 
                >{\centering\arraybackslash}p{0.22\textwidth} 
                >{\centering\arraybackslash}p{0.22\textwidth}}
    \multicolumn{3}{c}{\footnotesize\textbf{Validation Result}}\\
    \multicolumn{3}{c}{\vspace{-2mm}}\\
    \tickicon & \crossicon & \crossicon \\
    \footnotesize Task description & \footnotesize Test suite & \footnotesize Context relevance
\end{tabular}
    \end{subfigure}
\end{adjustbox}
\caption{Example of a contextualized programming task. This task has an \hlred{incorrect test} and fails to cover the \hlred{``Dictionaries''} concept. Our technique \textsc{PyTaskSyn} validates the quality of this task and abstains from outputting it to students.}

\label{fig.problemsetup.example}
\end{figure}

We define contextualized programming tasks in Section ~\ref{sec.problemsetup.contextualized}, introduce a quality metric in Section ~\ref{sec.problemsetup.qualityrubric} and technique evaluation metrics in Section ~\ref{sec.problemsetup.ourobjective}.

\subsection{Contextualized Programming Tasks}
\label{sec.problemsetup.contextualized}

\textbf{Context.} We define a context $\psi$ as a tuple $\psi = (\psi_{\theme}, \psi_{\concepts})$, where $\psi_{\theme}$ represents a theme of interest, and $\psi_{\concepts}$ denotes a set of target programming concepts for practicing while solving a task. The inclusion of such context aligns with prior works on generating contextualized programming tasks \cite{DBLP:conf/iticse/Gutierrez0L24,DBLP:conf/sigcse/Gutierrez0L24,DBLP:conf/icer/LogachevaHPS024}. 

\textbf{Programming task.} We define a programming task \T{} as being composed of a task description $\Tdesc$ and a test suite $\Ttests$. The task description $\Tdesc$ explains what needs to be accomplished, including the requirements, expected functionality, and constraints. The test suite $\Ttests$ is a set of test cases used to verify the correctness of a code with respect to $\Tdesc$. A student is given the task description $\Tdesc$ to solve, while the test suite $\Ttests$ is kept hidden to verify the student's code. A student solves task $\T$ successfully if their code passes all test cases in $\Ttests$. Figure \ref{fig.problemsetup.example} shows an example of a contextualized programming task.

\subsection{Quality of Contextualized Programming Tasks}
\label{sec.problemsetup.qualityrubric}
\looseness-1Ensuring the quality of a contextualized programming task is crucial before giving it to students. It must be relevant to the desired theme and programming concepts while being correct and comprehensible for students to solve. We propose a systematic evaluation process for a human expert to assess the quality of task \T{} created for context \(\psi\). The process begins with the expert formulating a solution for \(\mathcal{T}\), which is a natural approach to gain a thorough understanding of the task. If the expert cannot formulate a solution, this indicates a fundamental issue with the task, marking it as low-quality. Upon successfully formulating a solution code \(\mathcal{C}^*\), the expert evaluates the correctness of \(\mathcal{T}\) by verifying whether \Ttests{} correctly validates \(\mathcal{C}^*\), covering all base and corner cases handled by \(\mathcal{C}^*\). Next, they verify whether task \T{} is relevant to \(\psi_{\theme}\) and whether all the programming concepts in \(\psi_{\concepts}\) are required when formulating \(\mathcal{C}^*\). Then, they assess whether the task description \Tdesc{} provides sufficient information for students to write solutions. We define \emph{Q-Overall} as the overall quality metric, where an expert assigns a final numerical score at the end of the evaluation process.

\subsection{Technique Evaluation Metrics}
\label{sec.problemsetup.ourobjective}
To guide our technique development and evaluate its effectiveness, we use two performance metrics: (i) \emph{Coverage}, measuring the percentage of times a technique provides a programming task to the student, and (ii) \emph{Precision}, measuring the percentage of times the programming task provided to the student is of high quality. Our objective is to develop a technique with high precision to ensure well-designed tasks and high coverage rate to provide tasks across diverse contexts.

\section{Our Technique \textsc{PyTaskSyn}}
\label{sec.technique}

\looseness-01In this section, we first discuss the motivation and overview of our technique in Section~\ref{sec.technique.motivationandoverview}. Then, we detail each stage in our technique's pipeline, including task \emph{generation} in Section~\ref{sec.technique.generation} and task \emph{validation} in Section~\ref{sec.technique.validation}.

\subsection{Motivation and Overview}
\label{sec.technique.motivationandoverview}

\begin{figure*}[t!]
\centering
	\includegraphics[width=0.95\textwidth,trim={6.5cm 0cm 4cm 2.5cm},clip]{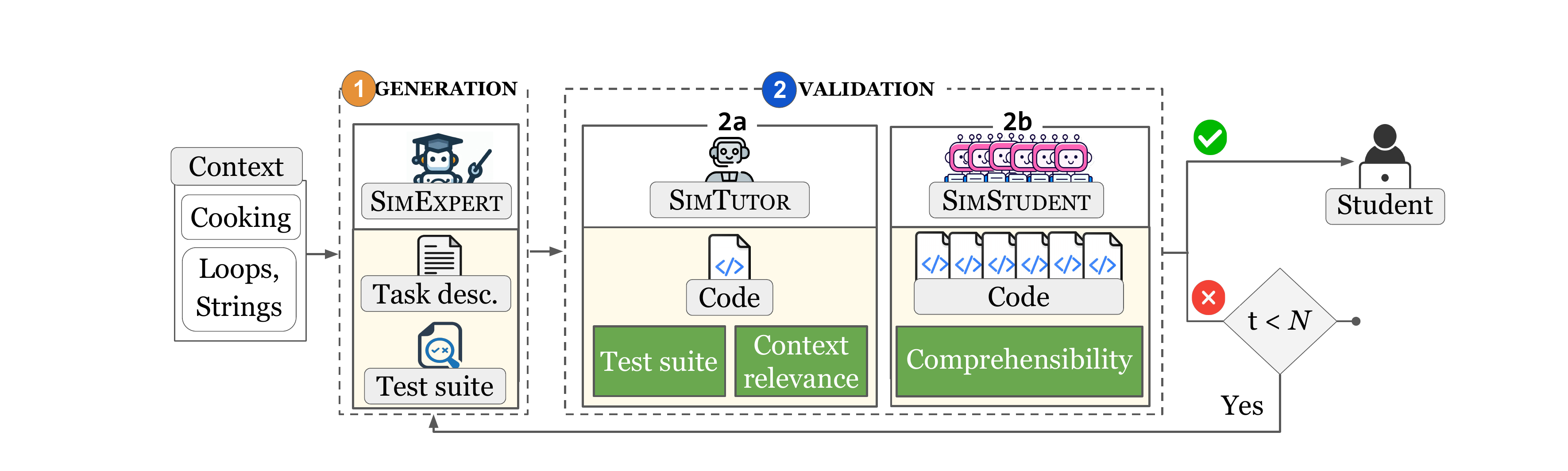}
    \vspace{-4mm}
    \caption{Pipeline of our technique \textsc{PyTaskSyn}. A task candidate is generated by a \textsc{SimExpert} agent given a theme and programming concepts (Stage \protect\orangecircled{1}). Then, \textsc{SimTutor} and \textsc{SimStudent} agents attempt to solve the task before assessing different quality aspects of the generated task candidate (Stage \protect\bluecircled{2} a-b). If the task does not pass the validation stage, our technique retries up to $N$ times, before ultimately deciding whether or not to assign a task to the student.}
    \label{fig.pipeline}
    \vspace{-2.5mm}
\end{figure*}

Common validation techniques use a single generative agent to generate both a task and its solution code for consistency checking \cite{DBLP:conf/icer/SarsaDH022} or to act as a judge for evaluating the output task \cite{DBLP:conf/nips/ZhengC00WZL0LXZ23}. However, our initial experiments revealed task quality issues when relying on a single agent (cf.\ Section~\ref{sec.evaluationresults}). Previous research show that breaking a complex process into sub-processes and combining the capabilities of multiple agents in a modular manner significantly improves performance \cite{DBLP:conf/iclr/ChenSZ0YCYLHQQC24,DBLP:journals/corr/abs-2308-08155}. Building on this insight, we implement a multi-agent technique where each agent takes a unique role. Figure~\ref{fig.pipeline} presents an overview of our technique consisting of a generation stage and a validation stage. The validation aims to improve precision, but may result in less coverage. To tackle coverage drop, we use an outer loop that repeats synthesis until a task passes validation or a maximum of $N$ trials is reached; if none pass, no task is given to the student.

\subsection{Stage 1 - Generation}
\label{sec.technique.generation} 

In this stage, we aim to generate a task \T{} composed of a task description \Tdesc{} and a test suite \Ttests{} for a given input context $\psi$. For this, our technique uses a simulated expert agent called \textsc{SimExpert}, implemented using the state-of-the-art generative model GPT-4o \cite{GPT4o}. Besides asking for the components of a task \T{}, we also request a solution code for \T{}, drawing inspiration from Chain-of-thought reasoning which has proven to enhance output quality from generative models \cite{DBLP:conf/nips/Wei0SBIXCLZ22}. We execute this code against \Ttests{} as a \emph{generation consistency} check \cite{DBLP:conf/icer/SarsaDH022}. If it does not pass \Ttests{}, task \T{} is considered invalid and will not go to the subsequent stages. Figure \ref{fig.prompts} (Stage 1) shows the structure of the prompt we use for this agent, with a system prompt establishing its role as a programming expert. The user prompt provides the task context $(\psi_{\theme}, \psi_{\concepts})$ and an instruction to generate a programming task. We use the model's default temperature of $1.0$.

\subsection{Stage 2 - Validation}
\label{sec.technique.validation} 

\textbf{Stage 2a - Validation by \textsc{SimTutor}}. In this stage, our goal is to evaluate the test suite and contextual relevance of the task generated in Stage 1. To this end, we implement a \textsc{SimTutor} agent to simulate a human tutor for validating the task. Figure~\ref{fig.prompts} (Stage 2a) shows the prompt we use for this \textsc{SimTutor} agent. It is assigned the role of a tutor through a system prompt, followed by a user prompt containing an input context $(\psi_\text{theme}, \psi_\text{concepts})$, and a generated task \T{}. First, we instruct the \textsc{SimTutor} agent to solve the task by writing a solution code $\mathcal{C}^*$. This approach follows the Chain-of-Thought prompting strategy \cite{DBLP:conf/nips/Wei0SBIXCLZ22}, which mimics the evaluating process of a human expert in Section \ref{sec.problemsetup.qualityrubric}. Solution $\mathcal{C}^*$ is then used to validate \Ttests{}. Specifically, we verify whether all test cases pass when executing $\mathcal{C}^*$ and whether every line of $\mathcal{C}^*$ is covered by \Ttests{}. Then, the agent assesses the relevance of the task to the given context. It assigns a score of 1 if task \T{} effectively integrates the given theme and programming concepts, and a score of 0 otherwise. We note that if the agent attempts to cheat by writing a code $\mathcal{C}^*$ that directly uses input-output pairs from \Ttests{}, the context relevance score will be 0 as it fails to use the required programming concepts. We use the GPT-4o model with its default temperature of 1.0 for our \textsc{SimTutor} agent.

\begin{figure*}[t!]
\centering
\footnotesize
\aboverulesep=0pt
\belowrulesep=0pt

\setlength\tabcolsep{0pt}
\renewcommand{\arraystretch}{1.2}
\setlength{\extrarowheight}{0pt}

\begin{tabular}{cp{0.92\textwidth}}
  \toprule
  
    \textbf{Stage} &
    \multicolumn{1}{c}{\textbf{Prompt template}} \\
    \toprule
  1 &
    \begin{tabular}[c]{@{}l@{}}
      \begin{lstlisting}[escapeinside={(*}{*)}]
  [System] You are an expert in Python programming.
  [User]     Given a theme of (*\textcolor{placeholder}{\{theme\}}*) and a list of programming concepts of
  (*\textcolor{placeholder}{\{concepts\}}*), generate a Python programming task that requires only the given
  programming concepts to solve. The task includes a description, a test 
  suite, and a solution program.
  \end{lstlisting}\end{tabular} 
    \\
    
    \hline
    
    2a &
    \begin{tabular}[c]{@{}l@{}}
    \begin{lstlisting}[escapeinside={(*}{*)}]       
  [System] You are a tutor in a Python programming course.
  [User]   The following Python programming task was created given a theme of
  (*\textcolor{placeholder}{\{theme\}}*) and a list of programming concepts (*\textcolor{placeholder}{\{concepts\}}*).
  Task description: (*\ \ \ \ \textcolor{placeholder}{\{task\_description\}}*)
  Test suite: (*\ \ \ \ \ \ \ \ \ \ \textcolor{placeholder}{\{testsuite\}}*)
  Write a program to solve the task and evaluate the context relevance of 
  the task. The context relevance is 1 if the task is clearly relevant to the 
  given theme and the theme is explicitly used throughout, and all given 
  programming concepts are strictly required to solve the task; 0 otherwise.
  \end{lstlisting}
    \end{tabular}
    \\
    \hline
  
  2b &
    \begin{tabular}[c]{@{}l@{}} \begin{lstlisting}[escapeinside={(*}{*)}]
  [System] You are a student enrolled in a Python programming course.
  [User]   Write a program to solve the task below.
  Task description: (*\ \ \ \ \textcolor{placeholder}{\{task\_description\}}*)
  
    \end{lstlisting}\end{tabular}
    \\
    \hline
  \bottomrule
  \end{tabular}

  \caption{
  \looseness-1 Overview of prompt templates for each stages implemented in \textsc{PyTaskSyn}. \textcolor{placeholder}{\{placeholders\}} are used to include details for concrete scenarios.}
  \label{fig.prompts}
\end{figure*}

\textbf{Stage 2b - Validation by \textsc{SimStudent}}. In this stage, we aim to evaluate the comprehensibility of the generated task \T{}. Specifically, the task description $\Tdesc$ should provide all the necessary information for a student to write a solution code. To this end, we use a classroom-scale population of \textsc{SimStudent} agents (20 agents in our experiments) to simulate students' points of view and obtain their solutions. If the majority of these simulated student agents fail to solve the task, $\Tdesc$ likely lacks clarity or critical information. We consider $\Tdesc$ comprehensible if at least $\tau$ percent (default at 50\%) of the \textsc{SimStudent} agents successfully solve task \T{} given only $\Tdesc$. To simulate students, we use a system prompt that assigns the agents the role of students, followed by a user prompt containing only the task description (see Figure \ref{fig.prompts}, Stage 2b). We use the GPT-4o-mini \cite{GPT4omini} model with a default temperature of 1.0, as prior research has shown that weaker models are better suited for simulating students' perspectives \cite{DBLP:conf/lak/PhungPS0CGSS24}.

\section{Evaluation}
\label{sec.evaluationsetup}
In this section, we present experimental evaluations centering around the following questions: (1) \textbf{RQ1}: How does \textsc{PyTaskSyn} perform compared to other existing techniques?; (2) \textbf{RQ2}: What are the contributions of different agents in \textsc{PyTaskSyn}?; and (3) \textbf{RQ3}: How does \textsc{PyTaskSyn} perform across different contexts? We present our evaluation setup in Section \ref{sec.evaluationsetup.contextselection}, \ref{sec.evaluationsetup.procedure}, \ref{sec.evaluationsetup.techniquesevaluated}, followed by results in Section \ref{sec.evaluationresults}. While our evaluations focus on the Python programming language, our technique can be extended to other programming languages.

\subsection{Context Selection} 
\label{sec.evaluationsetup.contextselection}
We evaluate the effectiveness of our technique across varied themes and programming concepts collected from prior works \cite{DBLP:conf/iticse/Gutierrez0L24,DBLP:conf/sigcse/Gutierrez0L24,DBLP:conf/icer/LogachevaHPS024,DBLP:conf/icer/SarsaDH022}. We select $5$ diverse themes and uniformly sample $5$ sets of $3$ to $5$ core Python programming concepts for each theme, resulting in $25$ contexts in total (cf. Figure \ref{fig.results.RQ3} for examples).

\subsection{Evaluation Procedure}
\label{sec.evaluationsetup.procedure}
For each sampled context $\psi$, we generate $N=10$ programming tasks, resulting in a pool of $250$ tasks ($25$ contexts $\times$ $10$ tasks). This is done prior to applying different validation techniques in Section \ref{sec.evaluationsetup.techniquesevaluated}. Two authors, with expertise in Python programming, evaluated the tasks using the Q-Overall metric introduced in Section \ref{sec.problemsetup.qualityrubric} and assigned binary scores (1-High quality/0-Low quality) to each task.\footnote{Additionally, our expert evaluation shows that 98\% of the 250 generated tasks do not contain programming concepts more advanced than those in the input context.} To better understand the reasoning behind evaluations, we ask the two annotators to answer three additional Yes(1)/No(0) questions: (i) Is the test suite correct and sufficiently covering relevant cases?; (ii) Does the task accurately reflect the input context?; (iii) Is the task description comprehensible? We obtain a Cohen's Kappa agreement \cite{Cohen1960ACO} of $0.8$ for Q-Overall metric and at least $0.7$ for each additional question, indicating substantial agreement between the annotators. We aggregate the two annotators' scores to obtain average quality scores for each task. Tasks passed validation of each technique in Section \ref{sec.evaluationsetup.techniquesevaluated} are used for computing precision and coverage.

\subsection{Techniques Evaluated}
\label{sec.evaluationsetup.techniquesevaluated}

\textbf{Baselines}. First, \GenBase{} neither applies a consistency check during the generation stage nor uses any validation mechanisms in the validation stage. Second, \GenConsistency{} incorporates only a consistency check during the generation stage \cite{DBLP:conf/icer/SarsaDH022}. Third, \LLMJudge{} leverages an LLM as a judge \cite{DBLP:conf/nips/ZhengC00WZL0LXZ23} to validate a task in the validation stage. We prompt it with input context $\psi$, a task \T{}, then instruct it to assess the test suite, contextual relevance, and task comprehensibility. It assigns a binary Q-Overall score (1/0) for the task's overall quality.

\textbf{Our Technique and Ablations}. $\PyTaskSyn$ involves consistency check in Stage 1 and multiple validation mechanisms in Stage 2. It can be parameterized by $\tau$, denoted as $\PyTaskSyn_{\tau}$, where $\tau$ represents the threshold for the percentage of \textsc{SimStudent} agents that solved the task. We implement two ablation variants, each corresponding to a stage of our validation pipeline. \SimTutorVal{} relies solely on validation from the simulated tutor agent, while \SimStudentVal{} uses only validation from simulated student agents. The default value of $\tau$ is set to $50\%$ for both \SimStudentVal{} and \PyTaskSyn{} in our evaluation.

\textbf{Oracle}. To evaluate the validation efficacy, we introduce $\Oracle_{p}$, which has access to ground-truth quality of tasks assessed by the human experts. It can select tasks to meet any precision threshold $p$, serving as an upper bound.

\begin{figure*}[t!]
    \centering
    \includegraphics[width=0.9\textwidth, trim={0 0.2cm 0 0cm}, clip]{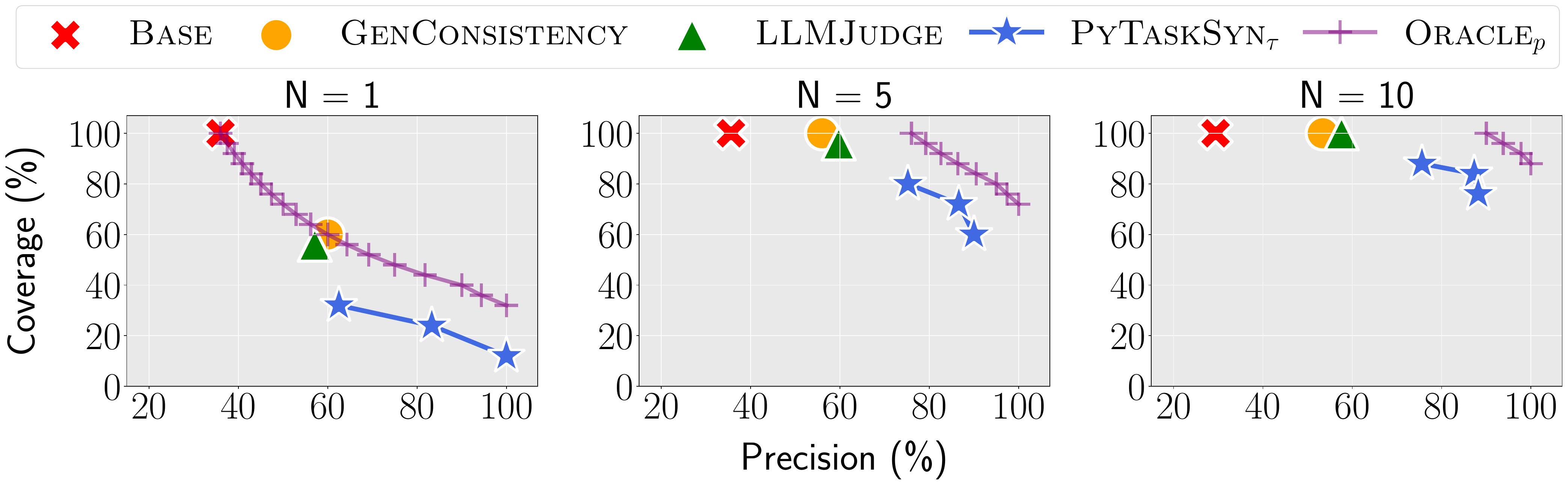}
    \caption{Technique comparison. $\PyTaskSyn_\tau$ improves task quality substantially over both baselines \GenConsistency{} and \LLMJudge{}, getting closer to the performance of $\Oracle_{p}$. We vary threshold $\tau\in\{0\%, 50\%, 100\%\}$ to showcase the tunable precision-coverage trade-off of $\PyTaskSyn_\tau$.}
    
    \label{fig.results.precisioncoverage}
\end{figure*}

\begin{figure*}[t!]
\centering
	\includegraphics[height=2.4cm, trim={0 2cm 0 1.8cm}, clip]{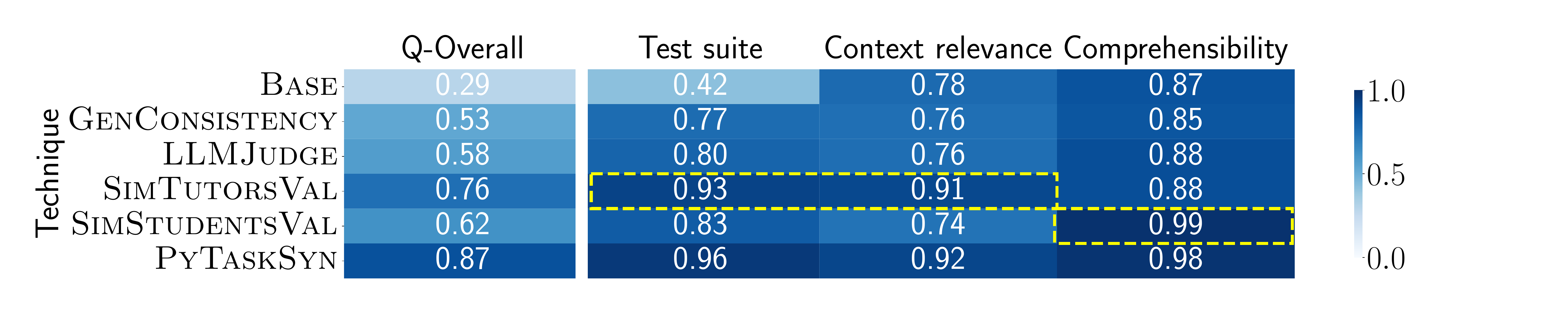}
    \caption{Ablation study. We report average Q-Overall scores and answers for three questions answered by experts (cf. Section \ref{sec.evaluationsetup.procedure}). Validation from the simulated tutor agent improves test suite quality and contextual relevance (as shown by \SimTutorVal{}), while validation from simulated student agents enhances task comprehensibility (as shown by \SimStudentVal{}). These agents collectively contribute to the overall quality of tasks, as demonstrated by \PyTaskSyn{}.}
    
    \label{fig.results.heatmap}
\end{figure*}

\subsection{Results}
\label{sec.evaluationresults}
\textbf{RQ1: How does \textsc{PyTaskSyn} perform compared to other techniques?} Figure \ref{fig.results.precisioncoverage} illustrates the precision and coverage of various techniques. Baseline methods including \GenBase{}, \GenConsistency{}, and \LLMJudge{} achieve high coverage but low precision $(\leq 60\%)$ due to no or insufficient validation. The subpar precision of \LLMJudge{} stems from a single agent's inability to thoroughly validate all aspects of task quality. Our technique $\PyTaskSyn_{\tau}$ leveraging perspectives from different simulated tutor and student agents improves the quality of synthesized tasks. As $N$ increases, $\PyTaskSyn_{\tau}$ not only increases coverage but also consistently achieves the highest precision compared to baselines. At $N=10$, \PyTaskSyn{} demonstrates strong performance, achieving a high precision of \( 87.3\% \) while maintaining substantial coverage at \( 84.0\% \). However, there is room for improvement in terms of both precision and coverage when compared to $\Oracle_{p}$. Finally, varying the passing threshold~$\tau$ in $\PyTaskSyn_{\tau}$ reveals a clear precision-coverage trade-off, allowing control based on requirements.

\textbf{RQ2: What are the contributions of different agents in \textsc{PyTaskSyn}?}
Figure~\ref{fig.results.heatmap} presents the ablation study analyzing the impact of different agent types on different task quality aspects. Our analysis reveals distinct and complementary contributions from the simulated agents. Simulated tutor agents enhance test suite quality notably and ensure strong context alignment (as shown by \SimTutorVal{}), while simulated student agents substantially boost task comprehensibility (as demonstrated by \SimStudentVal{}). Their combination in \PyTaskSyn{} achieves the highest overall quality, with Q-Overall score of $0.87$. This demonstrates how our multi-agent validation approach effectively combines different perspectives to ensure comprehensive task quality assessment.

\begin{figure*}[t!]
\centering
    \includegraphics[width=1\textwidth, trim={0 0.2cm 0 0}, clip]{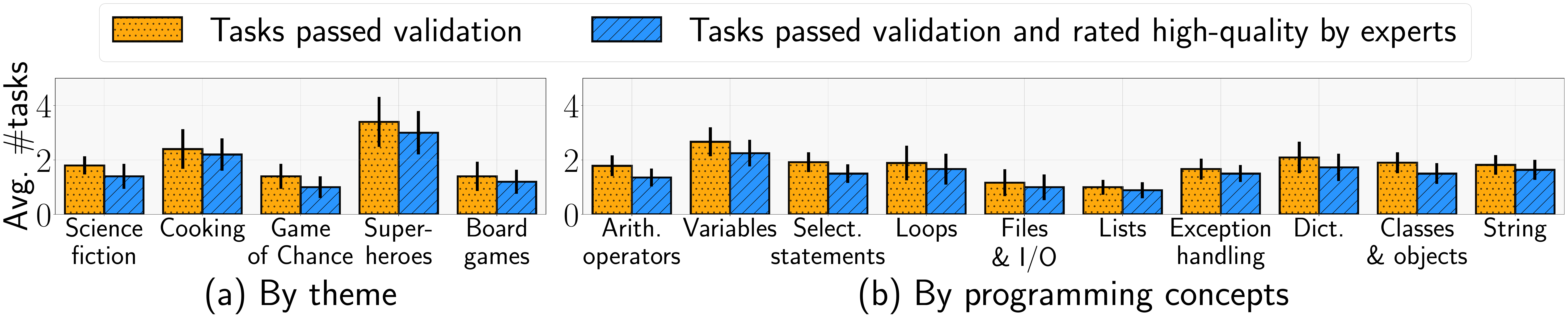}
    \caption{
    \looseness-1 Average number of tasks passed validation of \PyTaskSyn{} out of $N=10$ generated tasks, and average number of those rated as high-quality. \PyTaskSyn{} could deliver high-quality tasks for any theme and programming concept.}
    \label{fig.results.RQ3}
\end{figure*}
\textbf{RQ3: How does \textsc{PyTaskSyn} perform across different contexts?}
Figure~\ref{fig.results.RQ3} shows the average number of tasks synthesized by \PyTaskSyn{} as well as how many were rated as high-quality by experts across different themes and programming concepts. We observe that our technique performs well across most themes and programming concepts. Out of $N=10$ tasks, $1$ to $3$ tasks passed validation on average, with the majority rated as high-quality by human experts.


\section{User Studies Using Our Web Application}
\label{sec.pilotstudy}

We created a public web application (\url{https://pytasksyn.netlify.app}) where users can request tasks from \PyTaskSyn{} for their chosen themes and programming concepts, write code and debug using our integrated programming environment. We conducted the following two user studies via this web application.

\subsection{Comparison with Expert-created Tasks and Online Resources}
\label{sec.study1}

\textbf{Setup.} \looseness-1We compared tasks from \textsc{PyTaskSyn} against tasks created by experts and those from online resources. From 5 themes in Section~\ref{sec.evaluationsetup.contextselection}, we re-sampled $3$ to $5$ programming concepts to create 5 new contexts. For each, we collected $3$ tasks from: (1) an expert, (2) online resources (specifically geeksforgeeks.org), and (3) our web application.\footnote{We made our best effort to find tasks on geeksforgeeks.org that covered programming concepts in each context. The presence of specific themes generally cannot be found.} We recruited $10$ volunteer participants, including tutors and graduate students. They are non-native English speakers, with an average age of $28.3$. Among them, $5$ held Master's degrees and $5$ held Bachelor's degrees in STEM fields, with an average of $6.8$ years of Python experience. We assign each participant $2$ random contexts with their corresponding $6$ tasks (task sources are undisclosed). After completing each task or reaching the $20$-minute limit, they provided feedback using multi-level Likert scales on: theme relevance (1-Yes/0.5-Partial/0-No), programming concepts relevance (1-Yes/0.5-Partial/0-No), task comprehensibility (1-Yes/0-No), difficulty (1-Hard/0.5-Medium/0-Easy), and interestingness (1-Interesting/0.5-Okay/0-Boring).


\begin{figure}[t]
    \centering
        \footnotesize
    \setlength\tabcolsep{1.5pt}
    \aboverulesep=0pt
    \belowrulesep=0pt
    \renewcommand{\arraystretch}{1.1}
    \begin{tabular}{lcccccccc} 
    \toprule
                                                              & \begin{tabular}[c]{@{}c@{}}\textbf{Theme}\end{tabular} & \begin{tabular}[c]{@{}c@{}}\textbf{Concepts}\end{tabular} & \textbf{Compre.} & \textbf{Difficulty} & \textbf{Interest.} & \begin{tabular}[c]{@{}c@{}}\textbf{Success }\\\textbf{ (\%)}\end{tabular} & \begin{tabular}[c]{@{}c@{}}\textbf{Solving time}\\\textbf{(mins)}\end{tabular} & \begin{tabular}[c]{@{}c@{}}\textbf{Creation time}\\\textbf{(mins)}\end{tabular}  \\ 
    \hline
    Expert                                                  &$0.95$                                                                           &$0.95$                                       &$0.95$                                          &$0.42$                     &$0.72$                          &\phantom{1}$84$                                                                               &$10.82$                                                                                 &$25.20$                                                                                   \\ 
    Online resources                                           &$0.15$                                           &$0.82$                                      &$0.85$                     &$0.15$                          &$0.50$                              &$100$                                                 &$\phantom{1}3.26$                                                                                 & $0$\phantom{1}                                                                                   \\ 
    \textsc{PyTaskSyn}                       &$0.98$                                                                           &$0.92$                                        &$0.90$                                        &$0.20$                     &$0.65$                          &$100$                                                                               &$\phantom{1}6.66$                                                                                 &\phantom{1}$1.29$                                                                                   \\
    \bottomrule
    \end{tabular}
    \caption{Comparison of tasks from expert, online resources, and \textsc{PyTaskSyn}. We report averaged participant ratings on various task aspects and other statistics. Tasks from \textsc{PyTaskSyn} achieved quality comparable to expert-designed ones while requiring substantially less creation time.}
    \label{fig.study1}
\end{figure}
\begin{figure}[t]
\centering
    \scalebox{0.9}{
        \setlength\tabcolsep{4pt}
        \aboverulesep=0pt
        \belowrulesep=0pt
        \renewcommand{\arraystretch}{1.1}
        \begin{tabular}{lllllllllllc}
        \toprule

        \textbf{Statistic} &
        \multicolumn{10}{c}{\textbf{Participant ID}} &
        \textbf{Average} \\
        \cmidrule(lr){2-11}

        &
        P1 &
        P2 &
        P3 &
        P4 &
        P5 &
        P6 &
        P7 &
        P8 &
        P9 &
        P10 &
        \\
        
        \midrule
        No. Requests &
        $6$ &
        $5$ &
        $6$ &
        $6$ &
        $5$ &
        $5$ &
        $6$ &
        $5$ &
        $5$ &
        $5$ &
        $5.40$ \\

        No. Synthesized Tasks &
        $5$ &
        $5$ &
        $5$ &
        $5$ &
        $5$ &
        $5$ &
        $5$ &
        $5$ &
        $5$ &
        $5$ &
        $5.00$ \\

        No. Solved Tasks &
        $5$ &
        $5$ &
        $4$ &
        $4$ &
        $4$ &
        $5$ &
        $5$ &
        $5$ &
        $3$ &
        $5$ &
        $4.50$ \\

        Avg. Time to Solve (mins) &
        $4.94$ &
        $12.74$ &
        $12.66$ &
        $13.28$ &
        $5.46$ &
        $8.15$ &
        $10.14$ &
        $6.92$ &
        $7.50$ &
        $6.69$ &
        $8.85$ \\

        \bottomrule
        \end{tabular}
        }
    \caption{Statistics of each participant when they were instructed to request and solve tasks synthesized by \PyTaskSyn{} on our web application. Our system successfully synthesized tasks for 92.6\% of participant requests.}
    \label{fig.study2}
\end{figure}

\textbf{Results.} \looseness-1The multi-level feedback provided by participants is mapped to scores in the range $[0.0, 1.0]$, with mean scores for each source reported in Figure~\ref{fig.study1}. Additionally, we compute averaged participant success rate, solving time, and time taken to get a task from each source. Our analysis revealed that tasks synthesized by \textsc{PyTaskSyn} achieved quality comparable to expert-designed tasks while requiring significantly less creation time. Moreover, \textsc{PyTaskSyn} has an average task creation cost of just $0.13$ USD (APIs cost). Tasks from online resources, while readily available, generally lacked thematic integration and were rated as less interesting and easy. Expert-created tasks are the most challenging, resulting in longer solving times and lower success rates among participants.

\subsection{Real-world Performance of Our Web Application}
\label{sec.study2}
\textbf{Setup.} To evaluate our web application's real-world performance, we conducted a follow-up study with the same 10 participants. Unlike the previous study where participants solved pre-generated tasks from different sources, here each participant requested and solved 5 tasks in real-time through our web application. Participants selected their preferred contexts and made task requests until receiving a task. For each received task, we maintained the same instructions, time limits, and feedback questions as in the previous study.

\textbf{Results.} Our web application successfully synthesized tasks for $50/54$ requests (92.6\% coverage). Figure~\ref{fig.study2} provides insights into each participant’s session. Participants managed to solve on average 90.0\% of the synthesized tasks, with an average solving time of $8.85$ minutes. When analyzing their feedback, synthesized tasks showed on average high alignment with chosen theme ($1.0$) and programming concepts ($0.95$), while maintaining good comprehensibility ($0.86$).


\section{Concluding Discussions}
We introduced \PyTaskSyn{}, a novel synthesis technique that leverages generative models as agents simulating different classroom roles to validate generated programming tasks. Each stage in our pipeline contributes uniquely to the validation process, collectively ensuring the creation of tasks that are of high-quality and meet diverse learning objectives. Through extensive expert evaluation and user studies, we demonstrated that our approach significantly improves the quality of generated programming tasks with minimal cost, while maintaining reasonable coverage across various themes and programming concepts. 

Our work brings two important implications for leveraging generative AI for computing education. First, we demonstrate how to effectively automate the quality assessment of generated programming tasks. This paves the way to reducing educator workload when designing practice exercises that target specific learning objectives in programming education. Second, our results show that by breaking down the task synthesis process into different stages, we can leverage generative models that simulate different agents for specialized roles. This agent-based approach opens up new opportunities for utilizing generative models to simulate learning analytics to generate and validate educational content.

Next, we discuss some limitations of our work and possible ways of approaching them in the future. First, we adopt an accept/reject approach during our validation; future work could employ a framework for refining programming tasks by leveraging feedback from the validation stage. Second, we do not analyze whether the passing threshold of simulated students affects the synthesized task's difficulty; it would be valuable to investigate it and see whether it aligns with difficulty assessed by experts. Third, we focus on Python programming; it would be interesting to explore whether the capabilities of generative models for validating tasks extend beyond Python. Fourth, we conducted studies with a relatively small number of participants; it would be important to conduct larger-scale studies with students and assess their learning outcomes.

\begin{ack}
Funded/Co-funded by the European Union (ERC, TOPS, 101039090). Views and opinions expressed are however those of the author(s) only and do not necessarily reflect those of the European Union or the European Research Council. Neither the European Union nor the granting authority can be held responsible for them.
\end{ack}

\bibliographystyle{unsrt}
\bibliography{main}

\end{document}